\documentclass[letterpaper, 10 pt, conference]{ieeeconf}  

\IEEEoverridecommandlockouts                              

\usepackage{amsmath} 
\usepackage{amssymb}  
\usepackage{cases}
\usepackage{cite}
\usepackage{mathtools}
\usepackage{color,soul}
\usepackage{multirow}

\title{\LARGE \bf
Fly Safe: Aerial Swarm Robotics using Force Field Particle Swarm Optimisation
}

\author{Lauren Parker, James Butterworth and Shan Luo
\thanks{smARTLab, Department of Computer Science, University of Liverpool, Liverpool L69 3BX, UK. Emails: lauren.g.parker@outlook.com;
j.butterworth2@liverpool.ac.uk;
shan.luo@liverpool.ac.uk.}%
}

\usepackage[]{algorithm2e} 
\usepackage{amsmath,array}
\usepackage{graphicx}
\usepackage{subcaption}
\newcommand\Tstrut{\rule{0pt}{2.4ex}}       

\begin{document}

\maketitle
\thispagestyle{empty}
\pagestyle{empty}

\begin{abstract}

Particle Swarm Optimisation (PSO) is a powerful optimisation algorithm that can be used to locate global maxima in a search space. Recent interest in swarms of Micro Aerial Vehicles (MAVs) begs the question as to whether PSO can be used as a method to enable real robotic swarms to locate a target goal point. However, the original PSO algorithm does not take into account collisions between particles during search. In this paper we propose a novel algorithm called Force Field Particle Swarm Optimisation (FFPSO) that designates repellent force fields to particles such that these fields provide an additional velocity component into the original PSO equations. We compare the performance of FFPSO with PSO and show that it has the ability to reduce the number of particle collisions during search to 0 whilst also being able to locate a target of interest in a similar amount of time. The scalability of the algorithm is also demonstrated via a set of experiments that considers how the number of crashes and the time taken to find the goal varies according to swarm size. Finally, we demonstrate the algorithms applicability on a swarm of real MAVs.


\end{abstract}




\begin{figure}
\centering
\begin{subfigure}[b]{0.5\textwidth}
   \centering
   \includegraphics[scale = 0.78]{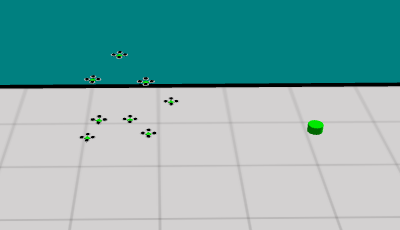}
   \caption{}
   \label{sim1} 
\end{subfigure}

\begin{subfigure}[b]{0.5\textwidth}
   \centering
   \includegraphics[scale = 0.78]{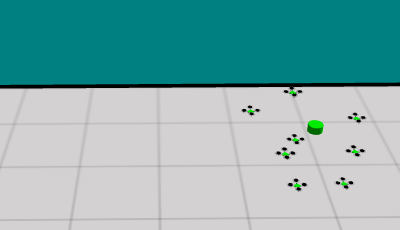}
   \caption{}
   \label{sim2}
\end{subfigure}

\caption[Two numerical solutions]{Snapshots of FFPSO in simulation. (a) shows the swarm move towards the goal location (the green puck) under the influence of FFPSO whilst avoiding collisions; (b) shows the swarm converge upon the goal location, illustrating the efficiency of FFPSO as a search procedure.}
\label{fig1}
\end{figure}

\section{INTRODUCTION}\label{intro}


Swarms in nature exhibit vastly complex behaviours in contrast to the simplistic reactive behaviours being carried out by the individuals. This is best illustrated in the way that termites build elaborate architectural structures or that ants forage for food via stigmergic processes. The fundamental property of a swarm, whether it be natural or artificial, is that the complex behaviour of the swarm is produced by the relatively simple behaviours carried out by the individuals - the product is by far greater than the sum of its parts.

There is a growing interest in using robotic swarms in real life problems thanks to a number of important properties they have over single robot systems: \textit{robustness} to individual failure, \textit{scalability} in terms of swarm size and environmental size and finally their ability to \textit{solve a problem in parallel} \cite{aerial_survey}. Furthermore, the falling costs of hardware and the improvements in communications, sensing and processing power are also driving the recent increase in interest \cite{aerial_survey}. Although the algorithmic and technological advancements have been significant in recent years, there are still a number of obstacles left standing between robotic swarms being used in scenarios such as search-and-rescue, area monitoring and the protection of safety critical infrastructure.



Particle Swarm Optimisation (PSO) is an optimisation procedure first introduced in order to visually represent the movement of a flock of birds \cite{pso}. It was soon discovered that making minor modifications to the simple flocking rules results in a very effective optimisation procedure. This procedure considers a number of ``particles'', each with a velocity, as points in a search space. This velocity is modified according to the personal best of the individual and global best of the swarm in relation to a fitness function. 
In this way, the movement of the particles across the search space is influenced by exploiting their own best known point and the best known point of the swarm. This information sharing via the global best allows the ``knowledge of the swarm'' to be accessed by the individual members. PSO has been successfully applied as a generalised optimisation procedure to various areas such as control systems and electrical engineering \cite{pso_survey}. 


Given that PSO is an effective search procedure, it seems natural to apply this concept to a swarm of real robots attempting to locate a goal. A swarm of this nature would be able to carry out safety critical tasks such as search-and-rescue, land mine detection and patrolling a border. However, one of the main problems with PSO, as highlighted in \cite{pso}, is that each particle is assumed to be an infinitesimally small, massless point; this is suitable for search over mathematical functions but not for use in real world robotics applications. Applying the original PSO algorithm directly to a robotic swarm would lead to individuals crashing into or coming dangerously close to one another as they move through the search space, especially in aerial swarms.

To make PSO applicable on aerial robotic swarms, we propose a novel algorithm named Force Field Particle Swarm Optimisation (FFPSO) that takes into account the repelling force field of each swarm member by adding an additional velocity component to the original PSO equation. 
Both the simulation tests (of which a snapshot is shown in Fig.~\ref{fig1}) and experiments on real Micro-Aerial Vehicles (MAVs) have shown that FFPSO is efficient at finding the location of a goal in 3-dimensional space using a swarm of aerial robots: it outperforms the original PSO algorithm in terms of convergence time and the number of crashes between individuals of the swarm.

\section{RELATED WORK} \label{relatedwork}

We begin by reviewing the most related work in the area of PSO, Potential Field methods and Flocking. PSO itself is a vast field with applications in many different areas (see \cite{pso_survey} for details), our aim here is not to cover the entirety of this but only what is relevant to aerial and swarm robotics. We also review some relevant work in the area of Potential Field methods, which are almost identical in nature to the ``force field" used in this work. However, we feel the alternative name is more appropriate in our work due to the 3-dimensional and finite nature of our fields acting around aerial robots. Finally, we review similar collision avoidance strategies employed in flocking algorithms.

\subsection{PSO in aerial robotic settings}
PSO has been applied to Unmanned Aerial Vehicles (UAVs) and MAVs in various ways already. Optimal route planning for MAVs is an optimisation problem that is tackled in \cite{uav_route_planning_1, Wang2015} by constructing complex fitness functions consisting of a number of different metrics that would affect the success of an MAV carrying out reconnaissance missions. 
These works modify the fitness function, whereas our work modifies the PSO equation directly. In the case of complicated fitness functions, the computational requirements of evaluating them for each individual at each time step could be far greater than our proposed method.
In \cite{pidpso,pidpso_2}, PSO is used to tune the parameters of a PID controller for an AR.Drone by constructing a multi-objective fitness function that takes into account a number of performance metrics w.r.t. the PID controller. 
In \cite{psoga}, PSO is hybridised with a Genetic Algorithm (GA) in order to optimise formation reconfiguration in swarms of UAVs. A hybrid algorithm is proposed that combines the advantages of both optimisation methods and is shown to outperform PSO in a series of simulated experiments. This algorithm optimises the control inputs of the UAVs such that optimal swarm reconfiguration can be achieved in battle-like simulations.

\subsection{Potential Field methods}

Potential Field (PF) methods are a set of algorithms that involve the simulation of artificial potential fields around objects or goals thus causing agents interacting in this environment to be repelled from objects and attracted to goals. PF methods have been applied in robotic motion planning \cite{Rimon1992, eapf, apf_1}, simulated swarms \cite{apf_swarm_1, apf_swarm_2, apf_swarm_3}, and more recently have been applied in real robotic swarms, including MAVs \cite{apf_swarm_real_1, apf_swarm_real_2}. In particular, in \cite{apf_swarm_real_1} PFs are applied to a swarm of MAVs under the remote control of a human operator. The operator controls the swarm as a single body and the potential fields generated by the robots and other objects in the environment help to prevent collisions, whilst keeping the formation of the swarm. 

In \cite{apf_swarm_real_2}, a centralised strategy for controlling a swarm of UAVs is devised for sowing seeds in a simulated field. Each individual is aware of the location of the seeds and sowing locations and moves in a straight line towards the target. The system employs a simple collision avoidance strategy that determines whether two drones are within a collision radius and if so, the agents move in the opposite direction of the collision. Despite not using an explicit potential field, the collision avoidance scheme uses similar vector operations to repel agents from collisions. In contrast with our work, when avoiding collisions the agents enter into a separate state, which interrupts the flow of the underlying search procedure and only accounts for one collision at a time. Our work amalgamates this collision vector with the vector of the trajectory to the goal (PSO velocity), thereby removing the need for a finite state machine and the separation of the collision avoidance state and the target locating state. We provide evidence for the fact that this state switching collision avoidance mechanism leads to a slower convergence speed in Section \ref{results_sim}. Although \cite{apf_swarm_real_1} and \cite{apf_swarm_real_2} employ these methods on real MAV swarms, they do not use PSO. Furthermore, both of these works only consider collision avoidance in 2 dimensions. 




The work most related to ours in theoretical approach is \cite{pugh2007}. In this work each individual ePuck robot represents a particle in the PSO algorithm where the aim is to find an area of interest. However, the main contributions of our work compared to \cite{pugh2007} is that we extend this model to 3 dimensions for aerial vehicles and we show our algorithm operating on a real swarm, whereas \cite{pugh2007} only tests the algorithm in simulation. Similar to \cite{apf_swarm_real_2}, \cite{pugh2007} employs a simple Braitenburg collision avoidance scheme in which particles instantaneously move in opposite directions after a collision and then continue to follow the original velocity before the collision occurred. 


\subsection{Flocking}
The flocking algorithms that originally inspired PSO employ a variety of collision avoidance strategies that are worth noting. According to Reynolds, flocking can be accurately simulated according to three individual principles: short range repulsion, local velocity alignment and long range attraction to the flock center \cite{Reynolds}. Due to this, collision avoidance mechanisms must be implemented in order to adhere to the short range repulsion requirement. Some of the work on flocking implements collision avoidance using a linearly decreasing repulsion force between members of the flock \cite{vasarhelyi, viragh} and other work uses a non-linear force decrease \cite{sabine, turgut}. 
The flocking principles are often implemented by accumulating the vectors of the respective influences (short range repulsion, long range attraction etc.) into one direction vector. This is very similar to the PSO method, however, the environmental influences on the particle are different, for example there is no requirement for the particles to group together (in fact this would be highly detrimental to search). 

To the best of the authors' knowledge, force field like methods have not yet been combined with PSO in our proposed way nor applied to a swarm of real MAVs. With this work we aim to contribute to developing biologically inspired algorithms such that they are suitable, and useful, in real world robotic swarms. 

\section{METHODOLOGY} \label{methodology}


\subsection{Particle Swarm Optimisation}

Particle Swarm Optimisation \cite{pso_survey} aims to maximise a fitness function $f:\mathbb{R}^{n} \rightarrow \mathbb{R}$ by initialising a population of particles in an \textit{n}-dimensional search space\footnote{For the experiments in this paper Equation \ref{eq8} is used as a fitness function, however, any function can be used.}. The \textit{n}-dimensional position $\textbf{x}_{i}(t)$ and velocity $\textbf{v}_{i}(t)$ of particle $i$ at $t=0$ are initialised randomly. The velocity and position of particle $i$ are updated according to Equations \ref{eq1} and \ref{eq2} respectively. Equation \ref{eq1} calculates the new velocity at the next time step according to the current velocity $\textbf{v}_{i}(t)$, the current position $\textbf{x}_{i}(t)$, the personal best of the individual particle (according to the fitness function) $\textbf{p}_{i}(t)$ and the global best of the population $\textbf{g}(t)$. The values $r_{1}$ and $r_{2}$ are random numbers generated in order to add stochasticity to the algorithm. The tuneable hyperparameters $\omega$, $\theta_{1}$ and $\theta_{2}$ can be altered in order to change the influence of the respective components. Equation \ref{eq2} applies this new velocity $\textbf{v}_{i}(t+1)$ to the old position $\textbf{x}_{i}(t)$ in order to get the new position $\textbf{x}_{i}(t+1)$.

\begin{equation}
	\begin{split}
		\textbf{v}_{i}(t+1) = \omega \textbf{v}_{i}(t) +
		\theta_{1} r_{1}(\textbf{p}_{i}(t) - \textbf{x}_{i}(t)) \\ + \theta_{2} r_{2}(\textbf{g}(t) - \textbf{x}_{i}(t))
	\end{split}
	\label{eq1}
\end{equation}

\begin{equation}
		\textbf{x}_{i}(t+1) = \textbf{x}_{i}(t) + \textbf{v}_{i}(t+1)
	\label{eq2}
\end{equation}

Equation \ref{eq1} consists of three separate components that represent conceptually different ideas. The first component $\omega \textbf{v}_{i}(t)$ represents the inertia of the particle. Physical inertia is defined as the resistance of a physical object to a change in motion, therefore, in the case of PSO, the inertial component provides a resistance of its current velocity to the effects of the other components. The second component of the equation $\theta_{1} r_{1}(\textbf{p}_{i}(t) - \textbf{x}_{i}(t))$ represents the individual knowledge of the particle or the ``cognition'' part \cite{psoga} that pulls the particle in the direction of its best known position so far. The third component $\theta_{2} r_{2}(\textbf{g}(t) - \textbf{x}_{i}(t))$ is known as the ``social'' part that pulls the particle in the direction of the global best of the population. The interaction of these 3 components results in an algorithm that is relatively successful at finding global optima due to the fact that knowledge about the global best is shared by all particles in the search space. Each of these particles will be in different places in the search space which leads to a greater global knowledge about that space, as opposed to an individual point performing gradient descent.



\subsection{Force Field Particle Swarm Optimisation}

\begin{figure}
\centering
\includegraphics[scale = 0.65]{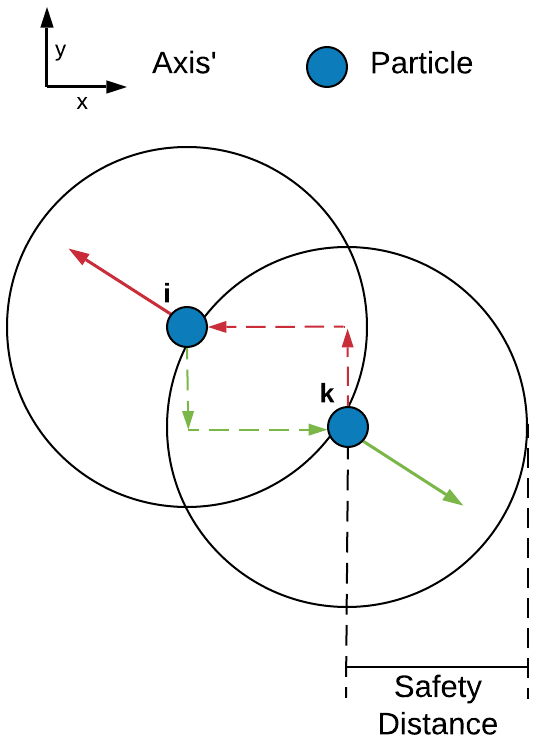} 
\caption{The force fields generated by the particles $i$ and $k$ induce a weighted vector in the opposite direction of the other particles. When the particles are within safety distance of one another this additional force field component helps to steer them away from each other, whilst continuing to be under the influence of the other PSO components.}
\label{ffdiagram}
\end{figure}

The newly proposed FFPSO algorithm works by introducing an additional component into Equation \ref{eq1}. This component is a force field component and aims to affect the PSO velocity of the individual particle by taking into account the force fields of the other particles such that they are repelled from one another at close distances. Fig. \ref{ffdiagram} shows a 2 dimensional snapshot of the effect of these fields from a top-down viewpoint. Equation \ref{eq3} describes the new FFPSO equation for particle $i$:
\begin{equation}
	\begin{split}
		\textbf{v}_{i}(t+1) = \omega \textbf{v}_{i}(t) +
		\theta_{1} r_{1}(\textbf{p}_{i}(t) - \textbf{x}_{i}(t)) \\ + \theta_{2} r_{2}(\textbf{g}(t) - \textbf{x}_{i}(t))
		+ \theta_{3}\textbf{ff}_{i}(t),
	\end{split}
	\label{eq3}
\end{equation}
where $\textbf{ff}_{i}(t)$ is the sum of the fields from all the surrounding $N$ particles as in Equation \ref{eq4}, and $\theta_{3}$ is a tunable hyperparameter that determines the influence of the force field component in the overall PSO equation.

\begin{equation}
	\begin{split}
		\textbf{ff}_{i}(t) = \sum_{k=1, k \neq i}^{N} \textbf{ff}_{ik}(t). 
	\end{split}
	\label{eq4}
\end{equation}

The type of force field applied to each particle affects the performance of the algorithm - we consider two different field types. The first type of field is a linearly decreasing field (FFPSO-LIN), the force field component, $\textbf{ff}_{ik}(t)$, between particle $i$ and particle $k$ for this field is calculated as in Equation \ref{eq5}:
\begin{equation}
 \textbf{ff}_{ik}(t) =
  \begin{cases} 
   (S - \lVert \textbf{r}_{ik} \rVert)\hat{\textbf{r}}_{ik} & \text{if } \lVert \textbf{r}_{ik} \rVert \leq S \\
   \Vec{0}       & \text{if } \lVert \textbf{r}_{ik} \rVert > S 
  \end{cases}
  \label{eq5}
\end{equation}
where $ \textbf{r}_{ik} = \textbf{x}_{i}(t) - \textbf{x}_{k}(t)$; $ \lVert \textbf{r}_{ik} \rVert$ is the magnitude of the vector $ \textbf{r}_{ik} $, i.e., the distance between particle $i$ and particle $k$; $\hat{\textbf{r}}_{ik}$ is the unit vector between particles $i$ and $k$ defined in Equation \ref{eq6}; $S$ is the safety distance after which there is zero effect from the field.

\begin{equation}
	\begin{split}
		\hat{\textbf{r}}_{ik} = \dfrac{\textbf{x}_{i}(t) - \textbf{x}_{k}(t)}{\lVert \textbf{x}_{i}(t) - \textbf{x}_{k}(t) \rVert}
	\end{split}
	\label{eq6}
\end{equation}

The second type of field is inspired by Newton's Law of Gravitation in that the force field component is inversely proportional to the distance between the two particles (FFPSO-GRAV). Using this field, $\textbf{ff}_{ik}(t)$ is calculated as in Equation \ref{eq7} as opposed to the linear field in Equation \ref{eq5}:
\begin{equation}
 \textbf{ff}_{ik}(t) =
  \begin{cases} 
   \frac{1}{(\lVert \textbf{r}_{ik} \rVert - D_{ik}) ^{p}} \hat{\textbf{r}}_{ik} & \text{if } \lVert \textbf{r}_{ik} \rVert \leq S \\
   \Vec{0}       & \text{if } \lVert \textbf{r}_{ik} \rVert > S 
  \end{cases}
  \label{eq7}
\end{equation}
where $p$ is a variable that determines how \textit{quickly} the strength of the field decreases with distance, and $D_{ik} = R_{i} + R_{k}$ where $R_{i}$ and $R_{k}$ are the respective radii of particle $i$ and particle $k$. $D_{ik}$ is mainly used to account for large robot diameters that need to be considered in order to avoid collisions. Furthermore, the separation of $D_{ik}$ into $R_{i}$ and $R_{k}$ accounts for particles of different radii. 

\begin{figure}
\centering
\includegraphics[scale = 0.25]{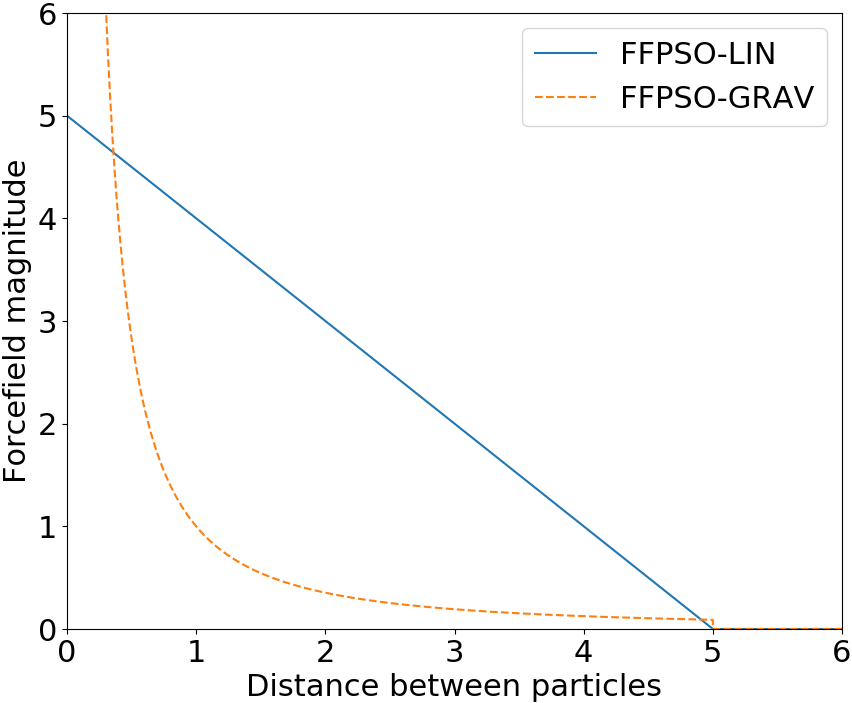} 
\caption{The strength of the force field w.r.t. the distance between 2 particles for both FFPSO-LIN and FFPSO-GRAV. The force field magnitude refers to the respective ``strength'' before the unit vector in Equations \ref{eq5} and \ref{eq7}. For FFPSO-LIN, this value is $S - \lVert \textbf{r}_{ik} \rVert$ and for FFPSO-GRAV this value is $ \frac{1}{(\lVert \textbf{r}_{ik} \rVert - D_{ik}) ^{p}}$. Constant values of $p=\frac{3}{2}, S=5$ and $D_{ik}=0$ are illustrated in this case.}
\label{ffgraph}
\end{figure}

To show the different effects of the two proposed force fields, Fig.~\ref{ffgraph} illustrates how the strength of the respective force fields change w.r.t. the distance between two particles. It shows that FFPSO-LIN provides a linear increase in force strength as particles approach one another. By contrast, FFPSO-GRAV induces a very small force near the safety distance which gradually increases as the particles come closer to each other. However, when the distance between the two particles is relatively small a dramatic increase in force strength ensues. The force strength for FFPSO-GRAV approaches infinity as the distance between the edges of the particles approaches 0, this theoretically prevents any crashes from occurring. The same cannot be said of FFPSO-LIN given that a maximum force strength of $S$ is achieved when the distance is 0, therefore if an opposing vector from the PSO equation is large enough it can override this force field resulting in a crash. Nonetheless, $S$ and $p$ can be altered in the equations to get different behaviours.

\section{EXPERIMENTAL SETUP} \label{experimentalsetup}

Experiments were conducted in the simulator ARGoS \cite{argos} and also in a physical lab on a real swarm of Crazyflie 2.0s (CF) from Bitcraze\footnote{https://www.bitcraze.io/crazyflie-2/}. The CF drones have a small size of 9 $cm^2$ and are very light, about 27 $grams$. 

\subsection{Simulation}\label{exp_setup_sim}

ARGoS is a multi-physics robot simulator that has been written in such a way that makes the simulation of large swarms highly efficient. The simulations are designed to be as close to real robots as possible, with very accurate models of real world robots already being available. For our simulations a generic quadrotor model was used, which is accurate enough to test the viability of the proposed algorithms and then transfer them to real CFs with minimal behavioural differences. 

The experiments were ran in a virtual arena of size $10m \times 10m \times 5m$ in order to simulate a real indoor environment in which a swarm of drones have to locate two goal locations one after the other. Only one member of the swarm is required to find the goal in order for the goal to be registered as found. At the beginning of each run, each agent was placed at a random position in the 3 dimensional space and assigned a random initial velocity. The goals are presented in the same position each time: the first goal is presented initially at position $(3, 5, 2.5)$ and then only after that goal is found by one member of the swarm the second goal is presented at position $(7, 5, 2.5)$. The fitness function at each point in space used for all of the algorithms is simply the negative of the distance between the particle and the goal as illustrated in Equation \ref{eq8}:
\begin{equation}
    f = - \lVert \textbf{g} - \textbf{x} \rVert
  \label{eq8}
\end{equation}
where $\textbf{g}$ and $\textbf{x}$ are 3-dimensional vectors representing the goal position and particle position respectively. After both goals have been found, or the maximum number of iterations has been reached which was 1,200, the simulation ends. After the simulation ends the number of iterations required to find both goals was recorded as well as the total number of crashes between agents.  

Given that PSO does not employ any collision avoidance scheme, we also compare an algorithm: PSO-CA, which is a modified version of PSO that does include a very simple collision avoidance mechanism. The collision avoidance scheme used is similar in nature to that used in \cite{pugh2007} and \cite{apf_swarm_real_2}: if two agents come within a safety distance of each other, the agents suspend their current computation and move in opposite directions to one another until out of the crash radius. Given that PSO has no explicit collision avoidance mechanism, we also compare the performance of our proposed methods against PSO-CA to further validate the effectiveness of our methods.

Each algorithm: PSO, PSO-CA, FFPSO-LIN (Equation \ref{eq5}) and FFPSO-GRAV (Equation \ref{eq7}) was tested on a swarm size ranging from 2 to 10 for 500 runs. The mean value of the number of iterations taken and the number of crashes were recorded for each set of runs. The original PSO hyperparameters used were $\omega = 1.0$, $\theta_{1} = 1.0$ and $\theta_{2} = 1.0$. For the FFPSO variants $\theta_{3} = 1.0$ was used and specifically for FFPSO-GRAV $p = \frac{3}{2}$ and $D_{ik}=0\ \forall i,k$ were used given that the diameter of the drones used in simulation were negligible. A safety distance of $0.4m$ was used for PSO-CA, FFPSO-LIN and FFPSO-GRAV. These values appeared most optimal according to a hyperparameter sweep. Fig. \ref{fig1} shows a snapshot of these experiments in the ARGoS simulator. 





\subsection{Real Robots}


\begin{figure}
\centering
\includegraphics[scale = 0.136]{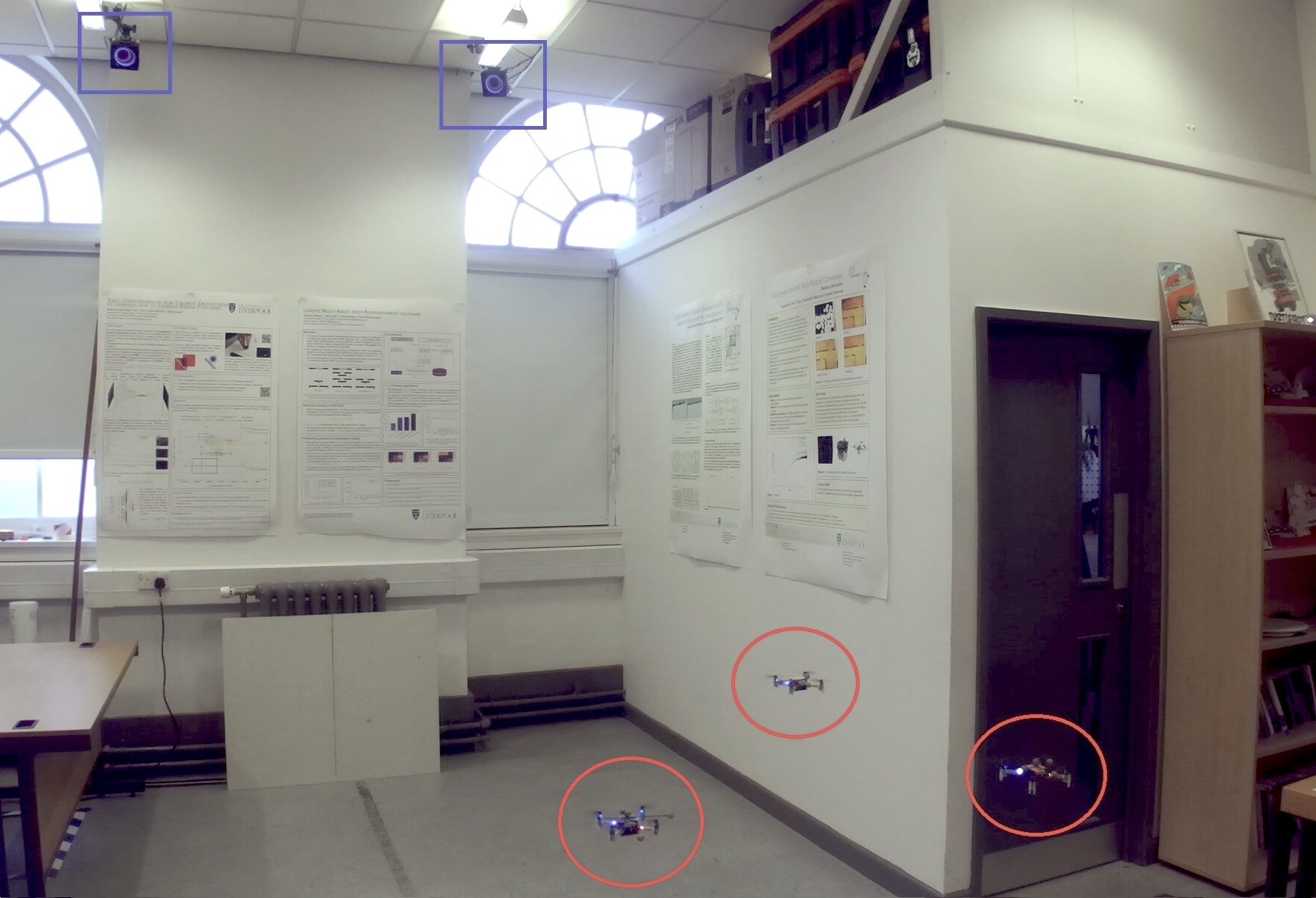} 
\caption{The real world setting. The motion capture cameras are identified by the blue squares and the CFs are identified by the red circles.}
\label{cfenv}
\end{figure}

The proposed algorithms were then implemented on a real swarm of three CFs, as shown in Fig. \ref{cfenv}. This was to show the suitability of the algorithm on a swarm of real drones and to consider any potential problems in a real robotic setting. Furthermore, we consider how the performance of the algorithm w.r.t. the time taken to find the goal is affected by transference from simulation to a real system. The quadcopters are stabilised using a Motion Capture System and a low level PID controller that accepts velocity commands as input and accelerates the CF to the desired speed. 

The CFs were placed randomly in an arena of size $3.9m \times 1.8m \times 2.5m$ and took off to a height of $1m$ after which the algorithm was started. The two goals were set $2m$ apart. The hyperparameters were adjusted for these experiments such that: $\omega = 0.6$, $\theta_{1} = 1.0$, $\theta_{2} = 1.0$ and $\theta_{3} = 1.0$. $\omega$ is reduced to stop the CFs velocities from being too large. The safety distance used in these experiments was $0.5m$. This experiment was ran 10 times and the time taken in seconds for the drones to move from the first goal to the second goal was recorded along with the flight path of each CF. During these experiments it was discovered that the downward draft from the CFs caused stability issues to CFs below it. Therefore, in the experiments the CFs were not allowed to fly beneath one another at a certain distance. This distance is known as the Motor Thrust distance and was set to 1.0m in our experiments; this is not integrated into the FFPSO-LIN equation it just completely disallows one CF flying beneath another CF.

\section{EXPERIMENTS \& ANALYSIS} \label{results}

In this section, the results for both the simulations and the real robotic experiments are given and analysed.

\subsection{Simulation}\label{results_sim}

Table \ref{crash_table} compares the mean number of crashes over 500 runs for each algorithm whilst also showing how this was affected by the number of agents in the swarm. It illustrates that having no collision avoidance scheme leads to a very high number of crashes, which is the case with PSO. The other three algorithms that do implement some form of collision avoidance either have 0 crashes over all of the experiments or have a very small number of crashes as is the case for FFPSO-LIN. 

The small number of crashes in FFPSO-LIN can be attributed to the fact that there exists a finite maximal force field strength when there is no space between the MAVs, which can be seen in Fig.~\ref{ffgraph}. Therefore, there exists a very small number of situations where this maximal value can be overridden by other components in Equation \ref{eq3}. The frequency of this occurrence was observed to be much higher with 2 drones due to the fact that each individual can attain a higher velocity. The configuration of a smaller swarm does not sufficiently hinder the momentum of the individuals such that crashes do not occur. Swarms of size 3 or more induce larger overall force field effects on one another which decelerates individuals more quickly, thereby preventing crashes. 

Nonetheless, Table \ref{crash_table} illustrates that the simple collision avoidance scheme added to PSO and our own force field schemes are very successful at eliminating crashes in a simulated swarm of agents carrying out the PSO search procedure. This is a highly desirable property that is essential for the application to a real swarm of MAVs.



\begin{table}[]
\begin{tabular}{llcccc}
 &  & \multicolumn{4}{c}{Algorithm} \\ \cline{2-6} 
\multicolumn{1}{l|}{\multirow{10}{*}{\rotatebox[origin=c]{90}{Number of Drones}}} & \multicolumn{1}{l|}{} & \multicolumn{1}{l|}{PSO} & \multicolumn{1}{l|}{PSO-CA} & \multicolumn{1}{l|}{FFPSO-LIN} & \multicolumn{1}{l|}{FFPSO-GRAV} \Tstrut\\ \cline{2-6} 
\multicolumn{1}{l|}{} & \multicolumn{1}{l|}{2} & \multicolumn{1}{c|}{1.704} & \multicolumn{1}{c|}{\textbf{0}} & \multicolumn{1}{c|}{2.274} & \multicolumn{1}{c|}{\textbf{0}} \\ \cline{2-6} 
\multicolumn{1}{l|}{} & \multicolumn{1}{l|}{3} & \multicolumn{1}{c|}{62.476} & \multicolumn{1}{c|}{\textbf{0}} & \multicolumn{1}{c|}{0.05} & \multicolumn{1}{c|}{\textbf{0}} \\ \cline{2-6} 
\multicolumn{1}{l|}{} & \multicolumn{1}{l|}{4} & \multicolumn{1}{c|}{134.758} & \multicolumn{1}{c|}{\textbf{0}} & \multicolumn{1}{c|}{0.044} & \multicolumn{1}{c|}{\textbf{0}} \\ \cline{2-6} 
\multicolumn{1}{l|}{} & \multicolumn{1}{l|}{5} & \multicolumn{1}{c|}{198.004} & \multicolumn{1}{c|}{\textbf{0}} & \multicolumn{1}{c|}{0.018} & \multicolumn{1}{c|}{\textbf{0}} \\ \cline{2-6} 
\multicolumn{1}{l|}{} & \multicolumn{1}{l|}{6} & \multicolumn{1}{c|}{256.418} & \multicolumn{1}{c|}{\textbf{0}} & \multicolumn{1}{c|}{0.028} & \multicolumn{1}{c|}{\textbf{0}} \\ \cline{2-6} 
\multicolumn{1}{l|}{} & \multicolumn{1}{l|}{7} & \multicolumn{1}{c|}{258.894} & \multicolumn{1}{c|}{\textbf{0}} & \multicolumn{1}{c|}{0.024} & \multicolumn{1}{c|}{\textbf{0}} \\ \cline{2-6} 
\multicolumn{1}{l|}{} & \multicolumn{1}{l|}{8} & \multicolumn{1}{c|}{224.106} & \multicolumn{1}{c|}{\textbf{0}} & \multicolumn{1}{c|}{0.022} & \multicolumn{1}{c|}{\textbf{0}} \\ \cline{2-6} 
\multicolumn{1}{l|}{} & \multicolumn{1}{l|}{9} & \multicolumn{1}{c|}{244.658} & \multicolumn{1}{c|}{\textbf{0}} & \multicolumn{1}{c|}{0.012} & \multicolumn{1}{c|}{\textbf{0}} \\ \cline{2-6} 
\multicolumn{1}{l|}{} & \multicolumn{1}{l|}{10} & \multicolumn{1}{c|}{234.614} & \multicolumn{1}{c|}{\textbf{0}} & \multicolumn{1}{c|}{0.038} & \multicolumn{1}{c|}{\textbf{0}} \\ \cline{2-6} 
\end{tabular}
\caption{The mean number of crashes between MAVs over a run of 1,200 time steps for each algorithm in simulation. 
PSO has a much larger number of crashes than any other algorithm given that it has no collision avoidance mechanism. Both PSO-CA and FFPSO-GRAV never result in any crashes whereas FFPSO-LIN only results in very few on average.}
\label{crash_table}
\end{table}


\begin{figure}
\centering
\includegraphics[scale = 0.15]{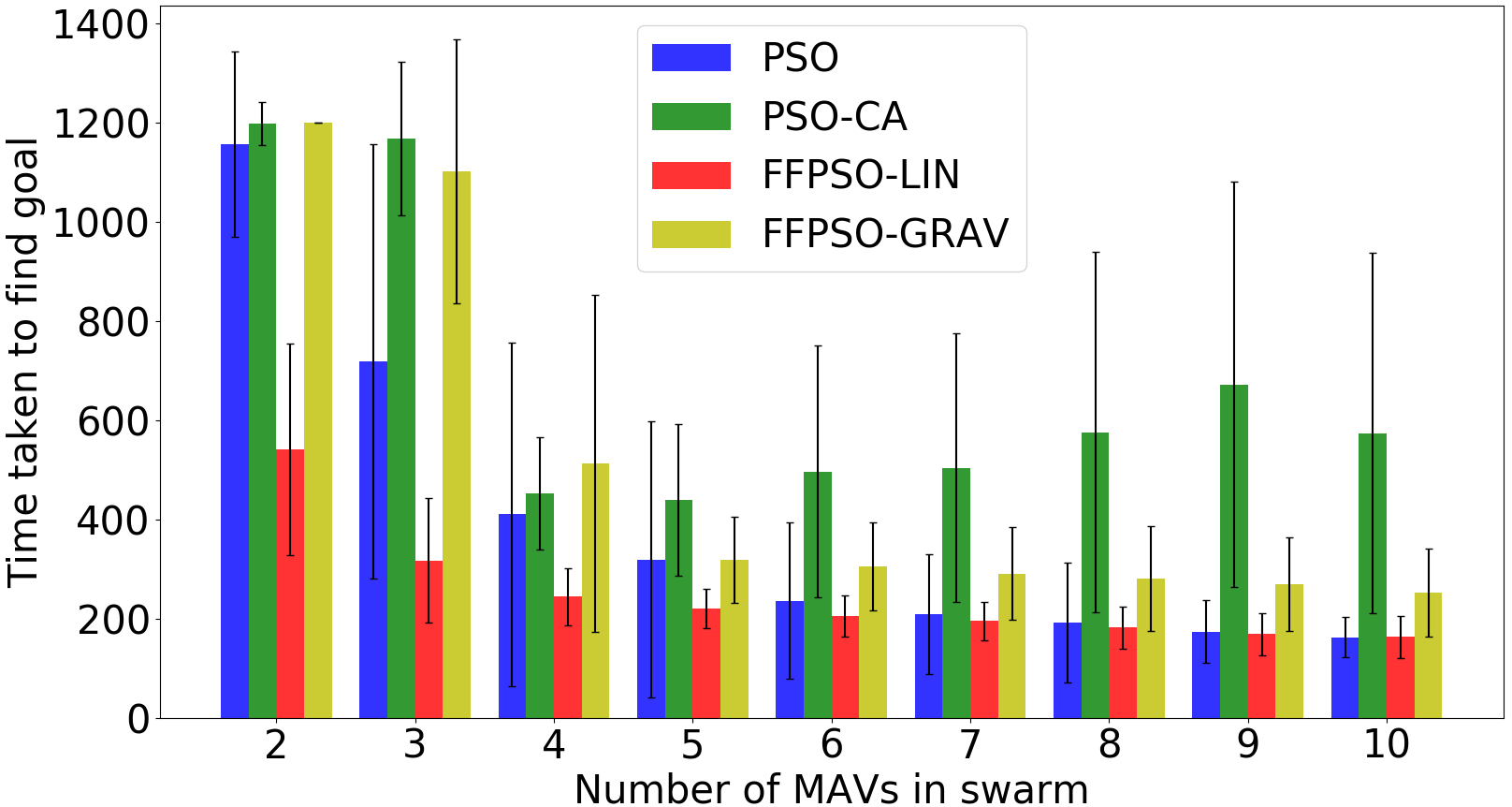} 
\caption{The number of time steps taken to find both goals w.r.t. the number of drones in the swarm. All algorithms except PSO-CA benefit from an increased number of drones in the swarm. FFPSO-LIN converges to the goal faster than any other algorithm and both force field based PSO methods outperform PSO-CA.}
\label{timevsnum}
\end{figure}



Fig.~\ref{timevsnum} shows how the number of time steps taken to find both goals changes w.r.t. swarm size for each of the algorithms tested. The general trend is that search time decreases as swarm size increases. This is expected from PSO and its variants whereby more information about the search space is generated by running more particles in parallel. Fig.~\ref{timevsnum} illustrates that FFPSO-LIN had a lower average time of completion than PSO-CA and FFPSO-GRAV for all swarm sizes and PSO for \textit{almost} all swarm sizes. FFPSO-LIN was 1.42 times faster on average over all swarm sizes compared to PSO. This can be attributed to the collision avoidance mechanism forcing the drones away from each other. This causes a larger swarm divergence and greater coverage across the search space leading to faster convergence to the goal than PSO without collision avoidance. This is most interesting to us and suggests that the inclusion of force fields into PSO does not hinder the convergence rate of the search in this domain. Furthermore, FFPSO-GRAV outperforms PSO-CA in all of the swarm sizes apart from one. Given that both these algorithms lead to no crashes at all, FFPSO-GRAV is the superior algorithm thanks to its increased convergence rate.


\subsection{Real Environment}

The experiments in the real environment were performed using the FFPSO-LIN algorithm on a swarm of 3 Crazyflie 2.0s. Due to battery limitations, the Crazyflie can only fly for a maximum of 1.5 minutes. As observed in Fig.~\ref{timevsnum}, it takes around 2 minutes (10 time steps for each second in the simulation) for FFPSO-GRAV to find both of the goals for a swarm size of 3. To this end, only FFPSO-LIN algorithm was implemented on the real swarm.

\begin{figure}
\centering
\includegraphics[scale = 0.5]{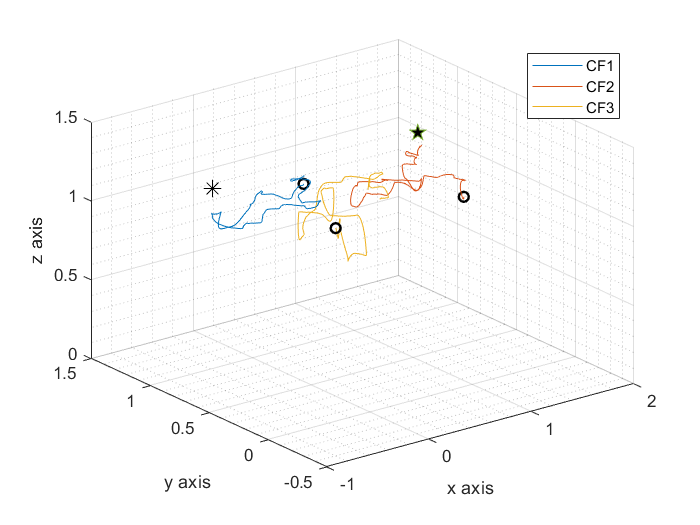} 
\caption{Flight paths of the three CFs on an example of a test flight.}
\label{paths}
\end{figure}

The time taken in seconds from locating the first goal ($\ast$ - in Fig. \ref{paths}) to locating the second goal ($\star$ - in Fig. \ref{paths}) was recorded for each run. The average time taken was 18.05 seconds over 10 runs which is a reasonably low amount of time for a swarm of 3 drones in this domain and there were no collisions. Fig. \ref{paths} shows the flight paths of each of the three CFs in one test. Each CF has their own coloured line representing their path and has a small black circle representing their starting position. As the CFs start searching CF1, shown as the blue line, and CF3, shown as the yellow line, avoid colliding and move away from each other. Later in the run CF1 locates the first goal as indicated by the blue line running near the first goal, then all the CFs change direction and travel in the direction of the second goal. As CF1 travels towards the second goal, CF3 has to avoid colliding with it as intended from the FFPSO algorithm. CF2, shown as the orange line, locates the second goal when it is within distance of detecting it. All of the CFs search for the goals while avoiding crashing into each other.

\begin{figure}
\centering
\begin{subfigure}[b]{0.5\textwidth}
   \centering
   \includegraphics[scale = 0.5]{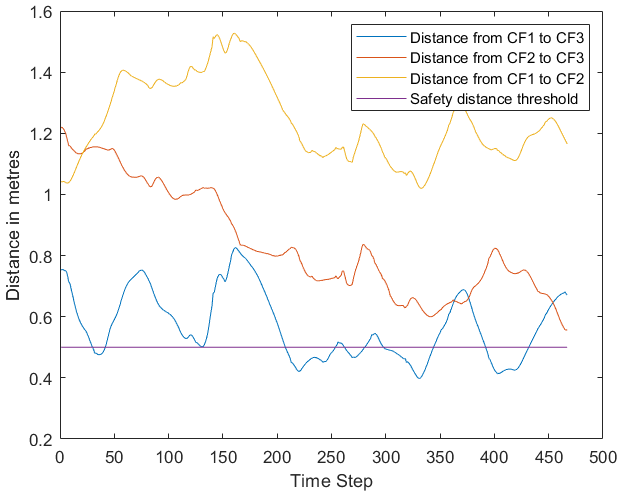}
   \caption{The distance between each of the individual members of the swarm.}
   \label{distance_to_cf} 
\end{subfigure}

\begin{subfigure}[b]{0.5\textwidth}
   \centering
   \includegraphics[scale = 0.5]{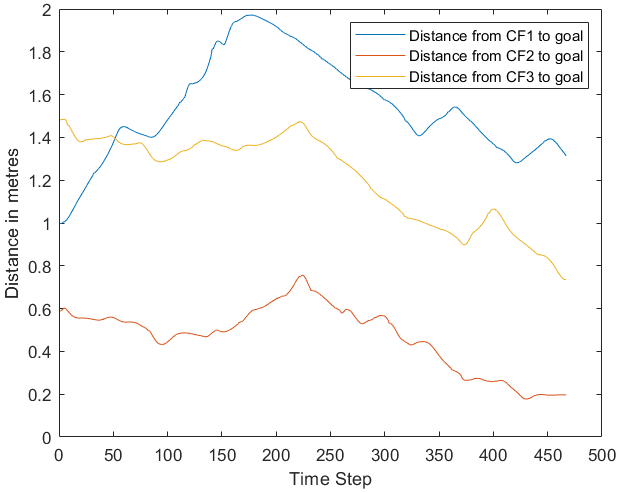}
   \caption{The distance between each of the members of the swarm and the second goal location.}
   \label{distance_to_goal}
\end{subfigure}

\caption{A couple of graphs showing the effectiveness of the collision avoidance mechanism and the search procedure of FFPSO-LIN during a real test flight.}
\label{real_distances}
\end{figure}

Fig.~\ref{distance_to_cf} shows how the distance between the CFs changes through time. The closest two CFs throughout the run are CF1 and CF3 as highlighted by the blue line. The graph shows that although the distance between these two MAVs does drop slightly below the safety threshold during the run, whenever it does, the force field component increases in strength and drives them apart again. Furthermore, Fig.~\ref{distance_to_goal} illustrates how the distance between the 3 members of the swarm and the second goal changes through time. For the first 200 time steps, all members move gradually towards the first goal (the opposite direction to the second goal). After this peak distance at approximately 200 time steps, all members of the swarm change direction and head towards the second goal, thereby reducing the distance between the members and the goal. Both of these figures illustrate both the collision avoidance and the PSO search procedure working effectively on a real MAV swarm.


\section{CONCLUSION \& FUTURE WORK} \label{conclusion}

In this paper, we propose a new algorithm named Force Field Particle Swarm Optimisation by combining force field methods with PSO. We provide a formal mathematical description of the modified PSO and we present and compare two different field types that are used as force fields. To test the performance of the newly proposed algorithm, it is compared in simulation on a varying size swarm of model MAVs to demonstrate its scalability. It is shown that FFPSO greatly reduces the number of crashes between particles to almost zero, whilst not affecting the search time in the case of FFPSO-LIN. The fact that FFPSO-LIN has a smaller search time than the original PSO method may highlight potential improvements that can be implemented in PSO as a generalised search procedure by continuing to impose a greater degree of coverage in the population throughout search; similar principles are employed in ideas such as Novelty Search \cite{NS}. We also demonstrate the applicability of our newly proposed algorithm to real aerial robotics via testing on a swarm of MAVs - this shows that the algorithm has the ability to be applied in a real world setting. We believe that our proposed force field methods will be a powerful tool in the future for providing simplistic collision avoidance mechanisms in the field of multi-agent aerial robotics.

Although the FFPSO implementation works well on a real swarm in this work, it is still dependent on a centralised server collecting personal bests, determining the global best and then passing this information to each individual in the swarm. This centralised approach relies on the server being in operation at all times and is less robust than a decentralised approach. Furthermore, it might not be applicable in real world scenarios where communication links with a central server might be severed due to adverse weather conditions, impenetrable materials or the individuals of a swarm drifting out of communication range. 

In future, we will look at converting our FFPSO algorithm implementation into a fully decentralised one in which the global best can be propagated to all individuals of the swarm without the use of a central server. 
We also plan on implementing our algorithms on swarms of greater size. Other MAV platforms such as Parrot drones that have longer flight time will be used to test the robustness of our algorithms on different platforms. This platform also has access to a high quality camera which we can use in order to perform tasks requiring vision such as search-and-rescue.


\section{Acknowledgements}
This work incorporates results from the research project ``Aerial Swarm Robotics for Active Inspection of Bridges'' funded by the Centre for Digital Built Britain (CDBB), under Innovate UK grant number 90066.






{
	\bibliographystyle{IEEEtran}
	\bibliography{ffpso.bib}
}

\end{document}